\pdfoutput=1

\documentclass[letterpaper, 10 pt, conference]{ieeeconf}  

\IEEEoverridecommandlockouts                              
\usepackage{amsmath} 
\usepackage{amssymb}  

\usepackage[pdftex]{graphicx}
\usepackage{graphics} 
\usepackage{cases}
\graphicspath{{./pic/}}
\DeclareGraphicsExtensions{.pdf,.jpeg,.png,.jpg}

\title{\LARGE \bf
3D Point-to-Keypoint Voting Network for 6D Pose Estimation
}

\author{Weitong Hua, Jiaxin Guo, Yue Wang and Rong Xiong
\thanks{Weitong Hua, Jiaxin Guo, Yue Wang and Rong Xiong are with the State key Laboratory of Industrial Control and Technology, Zhejiang University, Hangzhou, P.R. China. Rong Xiong is the corresponding author {\tt\small rxiong@iipc.zju.edu.cn}}
}

\begin{document}

\maketitle
\thispagestyle{empty}
\pagestyle{empty}

\begin{abstract}

Object 6D pose estimation is an important research topic in the field of computer vision due to its wide application requirements and the challenges brought by complexity and changes in the real-world. We think fully exploring the characteristics of spatial relationship between points will help to improve the pose estimation performance, especially in the scenes of background clutter and partial occlusion. But this information was usually ignored in previous work using RGB image or RGB-D data. In this paper, we propose a framework for 6D pose estimation from RGB-D data based on spatial structure characteristics of 3D keypoints. We adopt point-wise dense feature embedding to vote for 3D keypoints, which makes full use of the structure information of the rigid body. After the direction vectors pointing to the keypoints are predicted by CNN, we use RANSAC voting to calculate the coordinate of the 3D keypoints, then the pose transformation can be easily obtained by the least square method. In addition, a spatial dimension sampling strategy for points is employed, which makes the method achieve excellent performance on small training sets. The proposed method is verified on two benchmark datasets, LINEMOD and OCCLUSION LINEMOD. The experimental results show that our method outperforms the state-of-the-art approaches, achieves ADD(-S) accuracy of 98.7\% on LINEMOD dataset and 52.6\% on OCCLUSION LINEMOD dataset in real-time.

\end{abstract}

\section{INTRODUCTION}

Object 6D pose estimation is a task to calculate the rotation and translation of an object relative to the world coordinate system, with six degrees of freedom. It is an important topic for many applications, such as augmented reality \cite{marchand2015pose}, automatic driving \cite{xu2018pointfusion} and robot grasping \cite{zhu2014single,tremblay2018deep}. The challenges of this task come from the complexity in the real-world, such as background clutter, partial occlusion and illumination change. 

With the rapid development of deep learning on image, a lot of methods for pose estimation using RGB image have been proposed \cite{rad2017bb8,peng2019pvnet,zakharov2019dpod,park2019pix2pose}. Existing methods can be categorized as extended 2D object detection method \cite{kehl2017ssd,tekin2018real}, predicting pose directly by CNN \cite{xiang2017posecnn,do2018deep}, predicting 2D projection points \cite{rad2017bb8,peng2019pvnet} and template matching \cite{sundermeyer2018implicit}. But RGB image cannot provide rich feature information for industrial parts lacking texture information, and is vulnerable to illumination and other environmental changes. Some methods use only point cloud data to estimate object pose \cite{drost2010model}, which is more robust to light changes and texture-less object, but a large number of feature calculation and matching processes make the prediction very slow. In addition, it is not enough to use only geometric information of point cloud, while color and texture information are also important for pose estimation.

Some recent methods employ RGB-D data and focus on how to integrate the two kinds of data better to improve the accuracy of pose estimation \cite{xu2018pointfusion,li2018unified,wang2019densefusion}. For example, MCN \cite{li2018unified} considered the depth image as an additional channel of RGB image, then extracted the feature of this four-channel vector for pose prediction. PointFusion \cite{xu2018pointfusion} extracted the features of RGB image and point cloud respectively, and then directly concatenated them in the channel dimension. DenseFusion \cite{wang2019densefusion} proposed a novel point-wise way to fuse the features of RGB image and point cloud, which can better represent the object structure. But they all regress object translation and rotation directly, ignoring the spatial relationship information between points. Besides, the discontinuity of rotation makes the network difficult to learn.

\begin{figure}[t]
	\centering
	\includegraphics[width=1\linewidth]{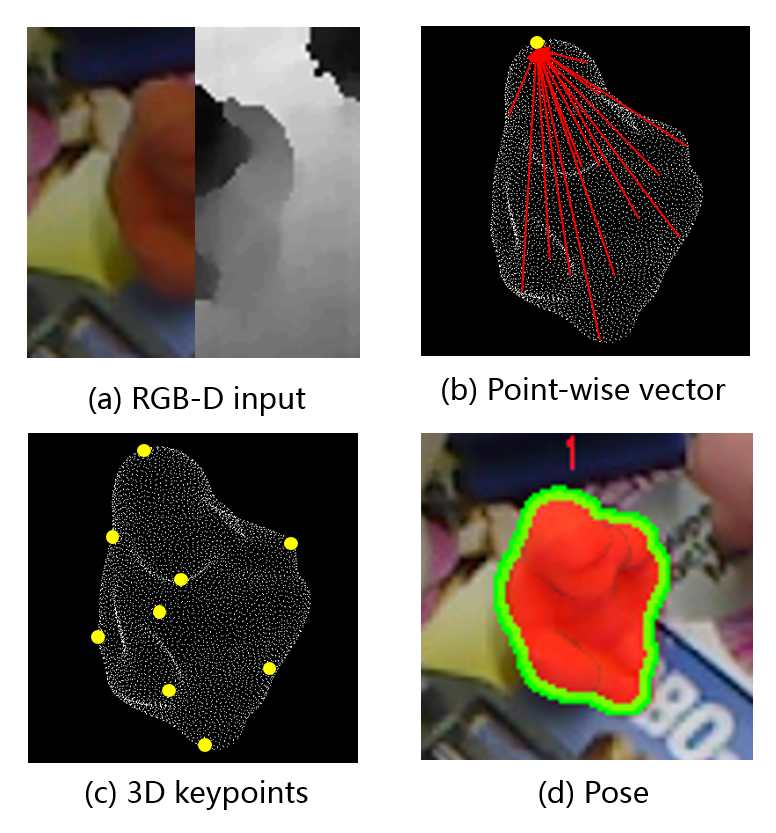}
	\caption{Several key intermediate results in our method are listed. In (a), we show the input RGB-D data. In (b), red lines denote direction vectors pointing to the yellow keypoint via network prediction. In (c), nine keypoints calculated by RANSAC voting is marked with yellow dots. In (d), the object model rendered according to the predicted pose is drawn on RGB image.}
	\label{fig:frame}
\end{figure}

To overcome these problems, we consider how to make full use of the spatial relationship between points, which is crucial for representation of object space state and is easy for CNN to optimize. However, using the spatial relationship of all points will involve a lot of work, resulting in low efficiency of calculation. So 3D keypoints are selected in order to extract the spatial relationship information, as shown in Fig. \ref{fig:frame}. This paper proposes a 6D pose estimation framework based on 3D keypoints, in which dense fusion feature is employed to vote for the keypoints. First, we extract features from RGB-D data individually and fuse them in the form of point-wise. Then the dense feature embedding is employed to predict direction vectors from sample points to 3D keypoints and the position of keypoints is inferred by RANSAC voting. Finally, pose can be calculated by the least square optimization algorithm. Our method can effectively learn the spatial structure information of the rigid body.

During implementation, there is an important problem that it is time-consuming to label the groundtruth of pose in the practical application. So some datasets have very limited training data, such as LINEMOD dataset \cite{hinterstoisser2011multimodal}, which may lead to overfitting problem. To avoid this problem, our network input is sampled in spatial dimension. During iterative training, partial point-wise embeddings are selected randomly to make the prediction instead of using all of them. In this way, the network input is different each time, which is equivalent to data augmentation, so our network can achieve good results even on small sample datasets.

In summary, this work has the following contributions:

\begin{enumerate}
	\item An object 6D pose estimation framework is proposed based on 3D point-to-keypoint voting. Point-to-point prediction enables our method to thoroughly learn the spatial structure characteristics of the rigid body, which can improve accuracy and efficiency. 
	
	\item During iterative training, the network input is sampled in spatial dimension to solve the overfitting problem under the small sample dataset, which can improve the generalization ability of the model.
	
	\item The performance on benchmark dataset outperforms the state-of-the-art method. The ADD(-s) accuracy is 98.7\% on LINEMOD dataset, and 52.6\% on OCCLUSION LINEMOD dataset. In addition, the time of pose estimation module is only 0.02s on a GTX 1060 GPU.
\end{enumerate}

The remainder of the paper is organized as follows: In Section II we briefly review related work on pose estimation. The proposed dense voting framework is introduced in Section III in detail. In Section IV we show the experimental results of our method on two benchmark datasets. Finally, a conclusion is given in Section V.

\section{RELATED WORK}

\textbf{Template matching methods.} Given CAD model of an object, some traditional methods extract different kinds of feature to match scene data and CAD model, in order to find the correct transformation. PPF \cite{drost2010model} proposes orientation point pair feature which is efficient and robust, then extracts it from model point cloud and scene point cloud respectively to match. Hinterstoisser \cite{hinterstoisser2011multimodal, hinterstoisser2012model} extracts contour gradient information and object surface normal vector from color image and depth image respectively as multimodal features for template matching. AAE \cite{sundermeyer2018implicit} collects RGB images of model from different perspectives and obtains many codebooks via autoencoder, then calculates the similarity with RGB images in the actual scene. Thanks to the CAD model of the object, these methods can easily acquire approximate results. But template matching always need to discretize the pose for comparing, which makes results inaccurate. 

\textbf{2D-3D correspondence methods.}  Inspired by object detection, some recent methods first find the 2D projection coordinates of keypoints on RGB image, and then employ 2D-3D correspondence to calculate pose by PnP \cite{zakharov2019dpod, park2019pix2pose, kehl2017ssd}. BB8 \cite{rad2017bb8} regards pose estimation as a 3D detection task and employs CNN to regress eight corners of 3D bounding box. YOLO-6D \cite{tekin2018real} extends 2D detection framework YOLO to 3D and also adopts 3D bounding box corners as keypoints. However, these bounding box corners are far away from the object pixels in the image, resulting in larger localization errors. To overcome this problem, PVNet \cite{peng2019pvnet} selects kepoints using the farthest point sampling (FPS) algorithm. Besides, instead of directly regressing the coordinate of the projection points, PVNet predicts the direction vector of each pixel pointing to the projection point, and then uses RANSAC based voting to get the position of the projection points. Pix2Pose \cite{park2019pix2pose} predicts the 3D coordinates of every pixel in RGB image and regards the output 3D coordinates as color image, thus transforms this problem into image generation problem based on GAN training. DPOP \cite{zakharov2019dpod} applies two-channel UV texture map and employs sphere projection or cylinder projection to add texture to the model, then predicts the dense 2D-3D mapping between RGB image and 3D model. These methods only use RGB image, which is convenient to collect, but all of them make no use of the important spatial relationship between points. 

\textbf{Regress pose methods.} With the development of deep learning in the field of computer vision, there are also some methods which employ CNN to extract features from RGB or RGB-D data and directly regress pose \cite{xiang2017posecnn, do2018deep, wang2019densefusion}. PoseCNN \cite{xiang2017posecnn} estimates translation and rotation respectively from RGB image via CNN. MCN \cite{li2018unified} further fuses the depth input as an additional channel of RGB image to a CNN architecture. PointFusion \cite{xu2018pointfusion} fuses them by directly concatenating the color feature with the geometry feature. These methods ignore the point cloud arrangement, which make the results ordinary. DenseFusion \cite{wang2019densefusion} presents a point-wise way to combine color and depth information from RGB-D input, which achieves good results. But these methods all regree pose directly, the nonlinearity of rotation space makes it difficult for CNN to optimize, limiting the learning ability of the model.

\begin{figure*}[thpb]
	\centering
	\includegraphics[width=1\linewidth]{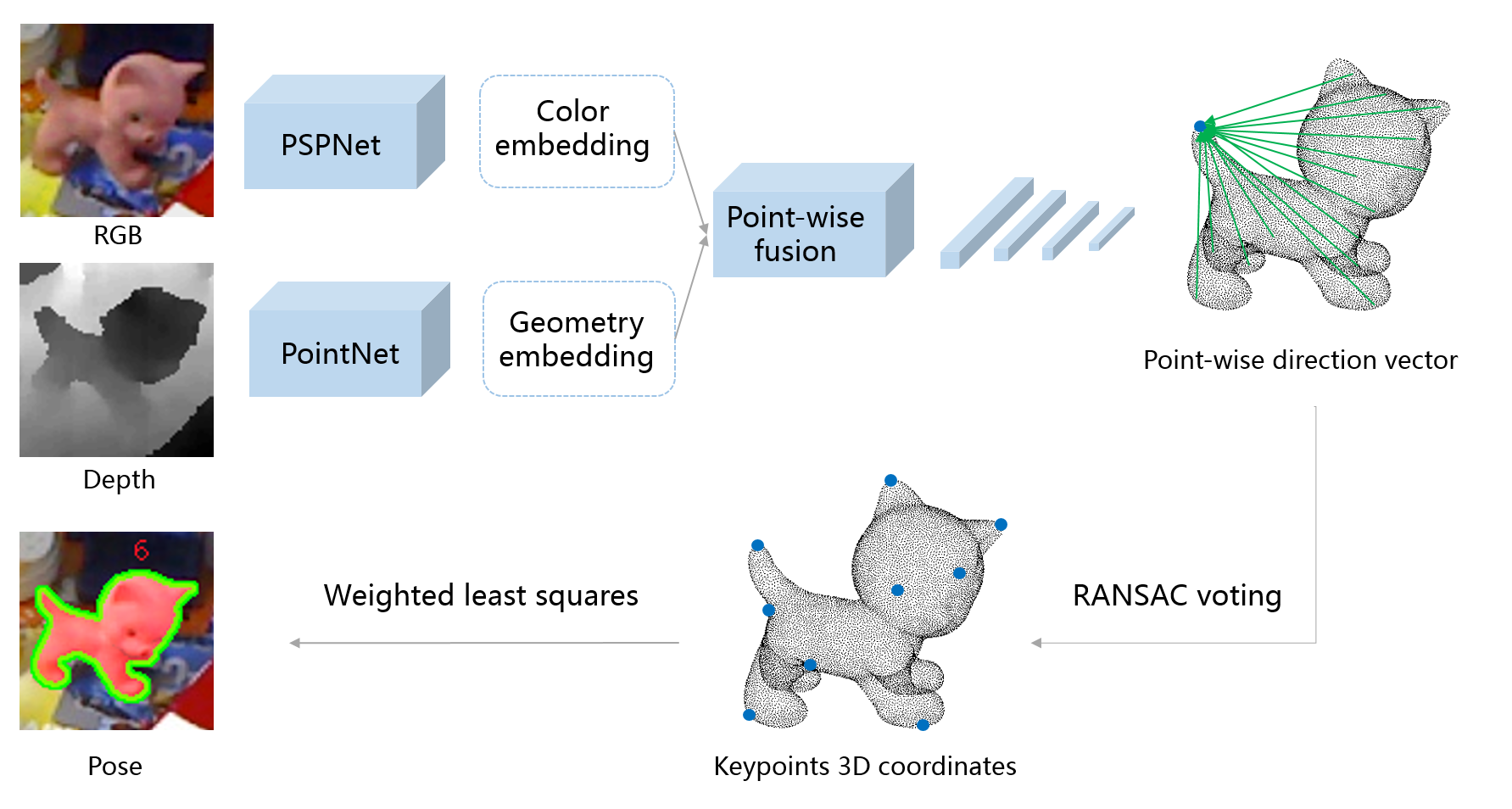}
	\caption{Overview of our 6D pose estimtion method: Given RGB image segmented from mask and point cloud transformed from depth image, the point-wise embedding is extracted to predict for direction vectors from sample points to 3D keypoints. The 3D corordinates of keypoints can be inferred by RANSAC based voting, and the final 6D pose can be obtained by the weighted least square optimization.}
	\label{fig:pipeline}
\end{figure*}

\section{METHOD}
Fig. \ref{fig:pipeline} shows the inference pipeline of our
proposing method which estimates the 6D object pose based on 3D keypoints. First, the features of RGB image and point cloud are extracted respectively via CNN, then dense fusion feature is obtained by point-wise way \cite{wang2019densefusion}. The following fully-connected layers are used to predict the direction vectors, which point to the 3D keypoints from spatial sampling points. Unlike regressing pose directly, the direction vector predicted by our network is easy to be optimized linearly. Besides, the point-to-point vector reflects the spatial relationship between points, which is important for accurate prediction of pose. Given a lot of vectors to each 3D keypoint, the coordinate of the 3D keypoints can be calculated based on RANSAC voting \cite{fischler1981random}. The final pose transformation is obtained by the weighted least square method, which aims to find the transformation matrix that makes the two sets of keypoints closest to each other.

\subsection{Point-wise Feature}

The feature fusion method in DenseFusion \cite{wang2019densefusion} is employed to obtain dense feature embedding. We need to segment the objects of interest in the
image before pose estimation. Then we employ network based on PSPNet \cite{zhao2017pyramid} for RGB image to extract color embedding, and employ network based on PointNet \cite{qi2017pointnet} for point cloud to extract geometry embedding. In order to obtain point-wise embedding, dense fusion procedure first concatenates the geometric embedding of each point and its corresponding color embedding. Then an average pooling layer is employed to generate a global feature vector, which is further concatenated to each
dense point-wise feature. In this way, the dense point-wise feature embedding we extracted contains both local and global information, which is more robust.

\subsection{Director Vector Prediction}

DenseFusion employs point-wise feature to regress translation and rotation directly. This framework has two shortcomings. On the one hand, rotation in 3D space is discontinuous, CNN can't learn the nonlinear value perfectly. On the other hand, this prediction way does not make full use of the spatial relationship among points.

Inspired by recent 2D methods \cite{rad2017bb8, peng2019pvnet}, we estimate pose by 3D keypoints instead of regression directly. Before training, FPS algorithm is employed to sample K 3D keypoints on CAD model surface like PVNet \cite{peng2019pvnet}. 

In order to let the network learn the spatial structure characteristics of point cloud, instead of regressing location of 3D keypoints directly, we predict the direction vector of each space point to the keypoints. Using many points to make predictions and then voting will be more stable than only making one prediction. For the space point $p$, the unit direction vector pointing to the keypoint $x_k$ is:

\begin{equation}
v_k = \frac{p-x_k}{||p-x_k||_2}
\label{eq:eq1}
\end{equation}

It should be noted that N scene points in spatial dimension are sampled during iterative training, so network will not overfit even training on small sample dataset, which can improve generalization ability.

The feature extraction network and direction vector prediction network are trained jointly, and the loss function is defined as the following equations.

\begin{equation}
Loss = \sum_{k=1}^K \sum_{p\in O}L(\Delta v_k(p)|_x)+L(\Delta v_k(p)|_y)+L(\Delta v_k(p)|_z), 
\end{equation}
\begin{equation}
\Delta v_k(p) = \widetilde v_k(p)-v_k(p),
\label{eq:eq2}
\end{equation}
where O is the set of N sampling points, $\widetilde v_k(p)$ is the predicted unit direction vector, $v_k(p)$ is the groundtruth value of unit direction vector. We decompose and calculate the smooth L1 loss in x, y and z directions:

\begin{equation}
L=smooth_{L1}(x) =
\left\{  
\begin{array}{lr}  
0.5x^2, &  if (|x|<1) \\  
|x|-0.5, &   otherwise
\end{array}  
\right.  
\label{eq:eq3}
\end{equation}

\subsection{3D Keypoints Voting}

After predicting a lot of direction vectors pointing to the keypoints by CNN, voting mechanism based on RANSAC \cite{fischler1981random} is employed to infer the coordinate of the 3D keypoints. Unlike 2D lines, two 3D lines may not be coplanar and there will be no intersection point for them. So we randomly select three points with their vectors each time, and employ the least square method to find the closest point to the group of vectors as one of the inferred positions $h_{k,i}$ of the keypoint $x_k$. The inferred position is computed as (5).

\begin{equation}
h_{k,i} = (\sum_i I-\hat v_i\hat v_i^T)^{-1}(\sum_i(I-\hat v_i\hat v_i^T)p_i),
\end{equation}
where $p_i$ is the matrix of the selected three points, and $v_i$ is the corresponding direction vector matrix.

The above process is repeated for M times to get the inferred location set, and then all the sample space points will vote for every inferred location by distance calculation. The number of vote $w_{k,i}$ can be seen as the confidence of each keypoint, which is computed in (6).

\begin{equation}
w_{k,i} = \sum_{p\in O}\mathbb I(\frac{(h_{k,i}-p)^T}{||h_{k,i}-p||_2}v_k(p)\geq\theta),
\end{equation}
where $O$ is the set of sample points, $\mathbb I$ denotes the indicator function, $\theta$ is a threshold which is assigned to 0.999 here.

The inferred position with the highest confidence is taken as the final position $s_k$ of the keypoint, and the corresponding confidence is $w_k$, see (7).

\begin{equation}
s_k = \arg\max_{h_{k,i}}w_{k,i}, \quad
w_k = \max w_{k,i},
\end{equation}

\subsection{Pose Calculate and Refine}

For K key points $s_k$ of the scene point cloud and K key points $m_k$ of the model point cloud, the weighted least square method is used to calculate the pose transformation, in which the weight is the confidence of each keypoint. The goal is to find the transformation matrix that makes these two keypoint sets closest to each other. The pose transformation matrix can be calculated as follows in (8).

\begin{equation}
R,t = \arg\min_{R,t}{\sum_{k=1}^Kw_k||(Rm_k+t)-s_k||^2}, 
\end{equation}

Finally, we employ the refine network proposed in DenseFusion \cite{wang2019densefusion} to iteratively optimize the pose: the scene point cloud is transformed according to the predicted pose and fed into the refine network to predict the pose residual. The final pose is determined by the predicted initial pose and residual, which is more accurate.

\section{EXPERIMENTS}
This section describes experimental results of our proposed method. We evaluate the performance and compare to the state-of-the-art methods on two datasets, LINEMOD dataset \cite{hinterstoisser2011multimodal} and OCCLUSION LINEMOD dataset \cite{brachmann2014learning}.

\begin{table*}
	\caption{ADD(-S) Accuracy on LINEMOD dataset.}
	\label{table:linemod}
	\begin{center}
		\begin{tabular}{c|cccccc|c}
			\hline
			object & BB8 & Pix2Pose & PVNet & PoseCNN & DenseFusion & DPOP & Ours\\
			\hline
			ape & 40.4 & 58.1 & 43.6 & 77.0 & 92.3 & 87.7 & \textbf{95.5}\\
			benchvise & 91.8 & 91.0 & \textbf{99.9} & 97.5 & 93.2 & 98.5 & 98.6\\
			cam & 55.7 & 60.9 & 86.9 & 93.5 & 94.4 & 96.1 & \textbf{99.3}\\
			can & 64.1 & 84.4 & 95.5 & 96.5 & 93.1 & \textbf{99.7} & 99.6\\
			cat & 62.6 & 65.0 & 79.3 & 82.1 & 96.5 & 94.7 & \textbf{99.4}\\
			driller & 74.4 & 76.3 & 96.4 & 95.0 & 87.0 & 98.8 & \textbf{99.1}\\
			duck & 44.3 & 43.8 & 52.6 & 77.7 & 92.3 & 86.3 & \textbf{95.3}\\
			eggbox & 57.8 & 96.8 & 99.2 & 97.1 & 99.8 & 99.9 & \textbf{100.0}\\
			glue & 41.2 & 79.4 & 95.7 & 99.4 & \textbf{100.0} & 96.8 & 99.8\\
			holepuncher & 67.2 & 74.8 & 81.9 & 52.8 & 86.9 & 87.7 & \textbf{98.6}\\
			iron & 84.7 & 83.4 & 98.9 & 98.3 & 97.0 & \textbf{100.0} & 99.8\\
			lamp & 76.5 & 82.0 & 99.3 & 97.5 & 95.3 & 96.8 & \textbf{99.3}\\
			phone & 54.0 & 45.0 & 92.4 & 87.7 & 92.8 & 94.7 & \textbf{98.8}\\
			\hline
			average & 62.7 & 72.4 & 86.3 & 88.6 & 94.3 & 95.2 & \textbf{98.7} \\
			\hline
		\end{tabular}
	\end{center}
\end{table*}

\begin{figure*}[thpb]
	\centering
	\includegraphics[width=1\linewidth]{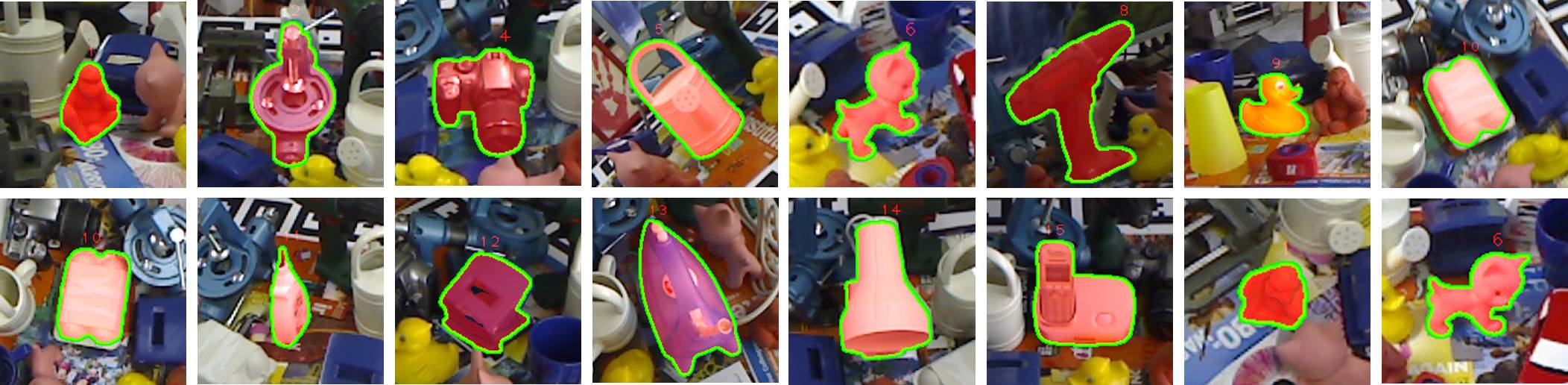}
	\caption{Some visualization results on LINEMOD dataset. In the pictures, the pose result is shown by projecting the model using the estimated pose. For the convenience of viewing, we also show the contour of the model projection.
	}
	\label{fig:linemod}
\end{figure*}

\begin{figure*}[thpb]
	\centering
	\includegraphics[width=1\linewidth]{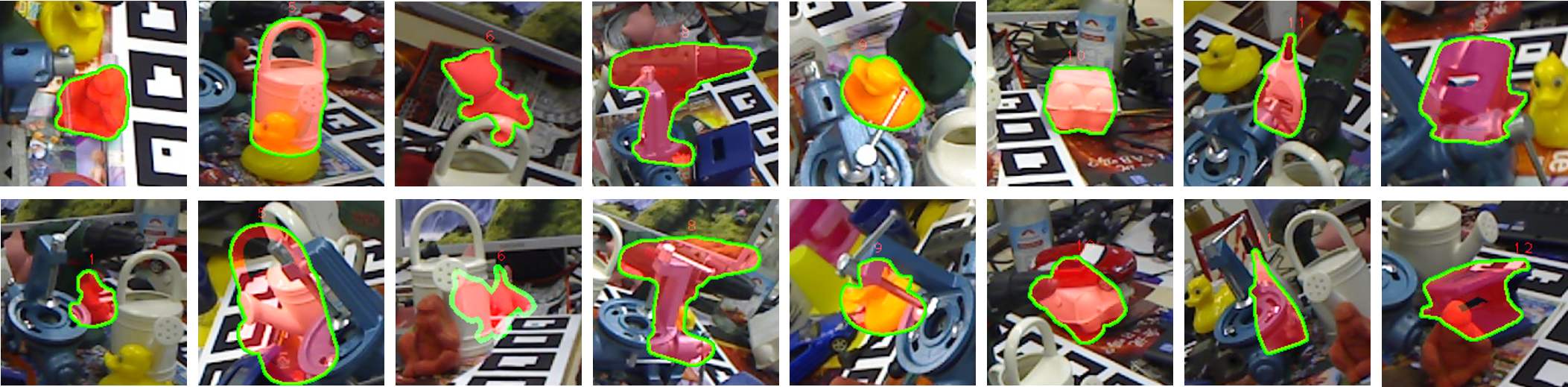}
	\caption{Some visualization results on OCCLUSION LINEMOD dataset. Pictures of the first line are some accurate results when the object is partially occluded, and pictures of the second line are some inaccurate results under heavy occlusion.
	}
	\label{fig:occ}
\end{figure*}

\begin{table}
	\caption{ADD(-S) Accuracy on LINEMOD dataset without tricks.}
	\label{table:trick}
	\begin{center}
		\begin{tabular}{c|ccc}
			\hline
			tricks & w/o refine & w/o fuse data & with both\\
			\hline
			accuracy & 94.8 & 96.6 & 98.7 \\
			\hline
		\end{tabular}
	\end{center}
\end{table}

\subsection{Datasets}
LINEMOD is a standard benchmark which is widely applied for 6D object pose estimation, so it is convenient to compare with other methods. This dataset contains 13 objects from different kinds. There are many challenges for pose estimation on LINEMOD dataset, such as low resolution, texture-less objects, cluttered scenes and lighting condition variations. 

OCCLUSION LINEMOD is a subset of the LINEMOD dataset by further annotation. It is only used for evaluation by model trained on LINEMOD dataset. Objects in this dataset have heavily occlusion, some of them even have a few pixels left to be seen. So it is a great challenge for pose estimation.

\subsection{Metrics}
We use ADD(-S) metric for the two datasets following prior works \cite{hinterstoisser2012model}. For non-symmetric objects, ADD metric is employed which is defined as average Euclidean distance between model points transformed with the predicted and the groundtruth pose respectively. The formula is as follows:

\begin{equation}
ADD = \frac 1 N\sum_{x\in O}||(Rx+t)-(\hat Rx+\hat t)||_2, 
\end{equation}
where $O$ is the set of model points, $N$ is the number of model points, $R$ and $t$
are the rotation and translation of groundtruth pose, $\hat R$ and $\hat t$ are the rotation and translation of predicted pose.

For symmetric objects (eggbox and glue in two datasets), due to the ambiguity of pose, ADD-S metric is employed which computes distance with the closest point. The formula is as follows:

\begin{equation}
ADD\text -S = \frac 1 N\sum_{x\in O}\min_{x_0\in O}||(Rx_0+t)-(\hat Rx+\hat t)||_2, 
\end{equation}

The evaluation result is deemed accurate when the average distance is less than 10\% of the object model’s diameter. 

\subsection{Implementation Details}

During implementation, 8 keypoints and 1 center point are selected for network to predict following PVNet. And during both training and testing, 500 scene points are randomly sampled as input. In order to enhance the robustness of illumination, online data augmentation such as light condition change is applied. In order to make the network insensitive to the background, we synthesize another 10000 images which randomly sample background from SUN397 dataset \cite{xiao2010sun} using the “Cut and Paste” strategy. These tricks are aim to avoid overfitting. The learning rate is setted as 0.0001 at first, and refine network is beginning to train with learning rate of 0.00003 when average distance error is less than 0.01.

\subsection{Results on benchmark dataset}

We report our quantitative evaluation results of the pose estimation experiments on the
LineMOD dataset (see Table \ref{table:linemod}). Our ADD(-S) accuracy is 98.7\%, which outperforms all other approaches. Ablation studies are also conducted to verify our performance without data augmentation and refine network (see Table \ref{table:trick}). Our network can aquire accuracy of 94.8\% without refine, while DenseFusion has only accuracy of 86.2\%. It confirms that our basic architecture is better than the state-of-the-art methods. Besides, the accuracy of our model which only trains on a small number of real data is 96.6\%, thus it can be seen that our spatial dimension sample strategy is helpful for avoiding overfitting. Some visualization results are shown in Fig. \ref{fig:linemod}, in which object CAD model is rendered on RGB image using estimated pose and the contour is also drawn on image.

As for OCCLUSION dataset, which is really difficult to estimate pose, we achieve ADD(-S) accuracy of 52.6\% and also outperforms other recent methods (see Table \ref{table:occ}). Some visualization results are shown in Fig. \ref{fig:occ}. Pictures of the first line show that our method can also estimate accurately when the object is partially occluded. But for some seriously occluded objects, our estimation is still biased, which can be seen in pictures of the second line.

Our pose estimation method only costs 0.02s on a GTX 1060 GPU, which is fast for real-time pose estimation.

\begin{table}
	\caption{ADD(-S) Accuracy on OCCLUSION LINEMOD dataset.}
	\label{table:occ}
	\begin{center}
		\begin{tabular}{c|cccc|c}
			\hline
			object & PoseCNN & Pix2Pose & PVNet & DPOP & Ours\\
			\hline
			ape & 9.6 & 22.0 & 15.8 & - & \textbf{51.6}\\
			can & 45.2 & 44.7 & 63.3 & - & \textbf{75.6}\\
			cat & 0.9 & 22.7 & 16.7 & - & \textbf{28.7}\\
			driller & 41.4 & 44.7 & 65.7 & - & \textbf{66.9}\\
			duck & 19.6 & 15.0 & 25.2 & - & \textbf{36.7}\\
			eggbox & 22.0 & 25.2 & \textbf{50.2} & - & 47.1\\
			glue & 38.5 & 32.4 & 49.6 & - & \textbf{71.9}\\
			holepuncher & 22.1 & \textbf{49.5} & 39.7 & - & 45.7\\
			\hline
			average & 24.9 & 32.0 & 40.8 & 47.3 & \textbf{52.6} \\
			\hline
		\end{tabular}
	\end{center}
\end{table}

\section{CONCLUSION}

An object 6D pose estimation method by 3D point-to-keypoint voting is proposed in this paper. Dense point-wise feature is employed to predict direction vectors, and 3D keypoints coordination is acquired by RANSAC voting. The final pose transformation is obtained by optimizing the distance between the scene keypoints and the model keypoints. During iterative training, a spatial dimension sample strategy is employed to solve the overfitting problem under the small sample dataset. Experimental results show that our method can effectively utilize the spatial structure information of the rigid body, which is better than the state-of-the-art methods in terms of accuracy and efficiency. 

In the future work, we plan to consider an one-stage method. We intend to predict confidence by CNN instead of RANSAC voting, and employ SVD to solve the least square method to make all processes derivable. Thus an end-to-end network can be designed which employs pose to supervise CNN learning to acquire more accurate results.


\addtolength{\textheight}{-3cm}   







\bibliographystyle{IEEEtran}
\bibliography{IEEEabrv,paper}

\begin{thebibliography}{10}
\providecommand{\url}[1]{#1}
\csname url@samestyle\endcsname
\providecommand{\newblock}{\relax}
\providecommand{\bibinfo}[2]{#2}
\providecommand{\BIBentrySTDinterwordspacing}{\spaceskip=0pt\relax}
\providecommand{\BIBentryALTinterwordstretchfactor}{4}
\providecommand{\BIBentryALTinterwordspacing}{\spaceskip=\fontdimen2\font plus
\BIBentryALTinterwordstretchfactor\fontdimen3\font minus
  \fontdimen4\font\relax}
\providecommand{\BIBforeignlanguage}[2]{{%
\expandafter\ifx\csname l@#1\endcsname\relax
\typeout{** WARNING: IEEEtran.bst: No hyphenation pattern has been}%
\typeout{** loaded for the language `#1'. Using the pattern for}%
\typeout{** the default language instead.}%
\else
\language=\csname l@#1\endcsname
\fi
#2}}
\providecommand{\BIBdecl}{\relax}
\BIBdecl

\bibitem{marchand2015pose}
E.~Marchand, H.~Uchiyama, and F.~Spindler, ``Pose estimation for augmented
  reality: a hands-on survey,'' \emph{IEEE transactions on visualization and
  computer graphics}, vol.~22, no.~12, pp. 2633--2651, 2015.

\bibitem{xu2018pointfusion}
D.~Xu, D.~Anguelov, and A.~Jain, ``Pointfusion: Deep sensor fusion for 3d
  bounding box estimation,'' in \emph{Proceedings of the IEEE Conference on
  Computer Vision and Pattern Recognition}, 2018, pp. 244--253.

\bibitem{zhu2014single}
M.~Zhu, K.~G. Derpanis, Y.~Yang, S.~Brahmbhatt, M.~Zhang, C.~Phillips,
  M.~Lecce, and K.~Daniilidis, ``Single image 3d object detection and pose
  estimation for grasping,'' in \emph{2014 IEEE International Conference on
  Robotics and Automation (ICRA)}.\hskip 1em plus 0.5em minus 0.4em\relax IEEE,
  2014, pp. 3936--3943.

\bibitem{tremblay2018deep}
J.~Tremblay, T.~To, B.~Sundaralingam, Y.~Xiang, D.~Fox, and S.~Birchfield,
  ``Deep object pose estimation for semantic robotic grasping of household
  objects,'' \emph{arXiv preprint arXiv:1809.10790}, 2018.

\bibitem{rad2017bb8}
M.~Rad and V.~Lepetit, ``Bb8: A scalable, accurate, robust to partial occlusion
  method for predicting the 3d poses of challenging objects without using
  depth,'' in \emph{Proceedings of the IEEE International Conference on
  Computer Vision}, 2017, pp. 3828--3836.

\bibitem{peng2019pvnet}
S.~Peng, Y.~Liu, Q.~Huang, X.~Zhou, and H.~Bao, ``Pvnet: Pixel-wise voting
  network for 6dof pose estimation,'' in \emph{Proceedings of the IEEE
  Conference on Computer Vision and Pattern Recognition}, 2019, pp. 4561--4570.

\bibitem{zakharov2019dpod}
S.~Zakharov, I.~Shugurov, and S.~Ilic, ``Dpod: Dense 6d pose object detector in
  rgb images,'' \emph{arXiv preprint arXiv:1902.11020}, 2019.

\bibitem{park2019pix2pose}
K.~Park, T.~Patten, and M.~Vincze, ``Pix2pose: Pixel-wise coordinate regression
  of objects for 6d pose estimation,'' in \emph{Proceedings of the IEEE
  International Conference on Computer Vision}, 2019, pp. 7668--7677.

\bibitem{kehl2017ssd}
W.~Kehl, F.~Manhardt, F.~Tombari, S.~Ilic, and N.~Navab, ``Ssd-6d: Making
  rgb-based 3d detection and 6d pose estimation great again,'' in
  \emph{Proceedings of the IEEE International Conference on Computer Vision},
  2017, pp. 1521--1529.

\bibitem{tekin2018real}
B.~Tekin, S.~N. Sinha, and P.~Fua, ``Real-time seamless single shot 6d object
  pose prediction,'' in \emph{Proceedings of the IEEE Conference on Computer
  Vision and Pattern Recognition}, 2018, pp. 292--301.

\bibitem{xiang2017posecnn}
Y.~Xiang, T.~Schmidt, V.~Narayanan, and D.~Fox, ``Posecnn: A convolutional
  neural network for 6d object pose estimation in cluttered scenes,''
  \emph{arXiv preprint arXiv:1711.00199}, 2017.

\bibitem{do2018deep}
T.-T. Do, M.~Cai, T.~Pham, and I.~Reid, ``Deep-6dpose: Recovering 6d object
  pose from a single rgb image,'' \emph{arXiv preprint arXiv:1802.10367}, 2018.

\bibitem{sundermeyer2018implicit}
M.~Sundermeyer, Z.-C. Marton, M.~Durner, M.~Brucker, and R.~Triebel, ``Implicit
  3d orientation learning for 6d object detection from rgb images,'' in
  \emph{Proceedings of the European Conference on Computer Vision (ECCV)},
  2018, pp. 699--715.

\bibitem{drost2010model}
B.~Drost, M.~Ulrich, N.~Navab, and S.~Ilic, ``Model globally, match locally:
  Efficient and robust 3d object recognition,'' in \emph{2010 IEEE computer
  society conference on computer vision and pattern recognition}.\hskip 1em
  plus 0.5em minus 0.4em\relax Ieee, 2010, pp. 998--1005.

\bibitem{li2018unified}
C.~Li, J.~Bai, and G.~D. Hager, ``A unified framework for multi-view
  multi-class object pose estimation,'' in \emph{Proceedings of the European
  Conference on Computer Vision (ECCV)}, 2018, pp. 254--269.

\bibitem{wang2019densefusion}
C.~Wang, D.~Xu, Y.~Zhu, R.~Mart{\'\i}n-Mart{\'\i}n, C.~Lu, L.~Fei-Fei, and
  S.~Savarese, ``Densefusion: 6d object pose estimation by iterative dense
  fusion,'' in \emph{Proceedings of the IEEE Conference on Computer Vision and
  Pattern Recognition}, 2019, pp. 3343--3352.

\bibitem{hinterstoisser2011multimodal}
S.~Hinterstoisser, S.~Holzer, C.~Cagniart, S.~Ilic, K.~Konolige, N.~Navab, and
  V.~Lepetit, ``Multimodal templates for real-time detection of texture-less
  objects in heavily cluttered scenes,'' in \emph{2011 international conference
  on computer vision}.\hskip 1em plus 0.5em minus 0.4em\relax IEEE, 2011, pp.
  858--865.

\bibitem{hinterstoisser2012model}
S.~Hinterstoisser, V.~Lepetit, S.~Ilic, S.~Holzer, G.~Bradski, K.~Konolige, and
  N.~Navab, ``Model based training, detection and pose estimation of
  texture-less 3d objects in heavily cluttered scenes,'' in \emph{Asian
  conference on computer vision}.\hskip 1em plus 0.5em minus 0.4em\relax
  Springer, 2012, pp. 548--562.

\bibitem{fischler1981random}
M.~A. Fischler and R.~C. Bolles, ``Random sample consensus: a paradigm for
  model fitting with applications to image analysis and automated
  cartography,'' \emph{Communications of the ACM}, vol.~24, no.~6, pp.
  381--395, 1981.

\bibitem{zhao2017pyramid}
H.~Zhao, J.~Shi, X.~Qi, X.~Wang, and J.~Jia, ``Pyramid scene parsing network,''
  in \emph{Proceedings of the IEEE conference on computer vision and pattern
  recognition}, 2017, pp. 2881--2890.

\bibitem{qi2017pointnet}
C.~R. Qi, H.~Su, K.~Mo, and L.~J. Guibas, ``Pointnet: Deep learning on point
  sets for 3d classification and segmentation,'' in \emph{Proceedings of the
  IEEE conference on computer vision and pattern recognition}, 2017, pp.
  652--660.

\bibitem{brachmann2014learning}
E.~Brachmann, A.~Krull, F.~Michel, S.~Gumhold, J.~Shotton, and C.~Rother,
  ``Learning 6d object pose estimation using 3d object coordinates,'' in
  \emph{European conference on computer vision}.\hskip 1em plus 0.5em minus
  0.4em\relax Springer, 2014, pp. 536--551.

\bibitem{xiao2010sun}
J.~Xiao, J.~Hays, K.~A. Ehinger, A.~Oliva, and A.~Torralba, ``Sun database:
  Large-scale scene recognition from abbey to zoo,'' in \emph{2010 IEEE
  Computer Society Conference on Computer Vision and Pattern
  Recognition}.\hskip 1em plus 0.5em minus 0.4em\relax IEEE, 2010, pp.
  3485--3492.

\end{thebibliography}

\end{document}